\pdfoutput=1
\def\year{2017}\relax
\documentclass[letterpaper]{article}

\usepackage{aaai17}
\usepackage{times}
\usepackage{helvet}
\usepackage{courier}

\usepackage{amsmath}
\usepackage{amsfonts}
\usepackage{graphicx}
\usepackage{multirow}
\graphicspath{{plots/}}
\usepackage[table,xcdraw]{xcolor}
\usepackage{siunitx}

\begin{document}
%

\title{Predicting Demographics of High-Resolution Geographies with Geotagged Tweets}
\author{Omar Montasser \and Daniel Kifer\\
School of Electrical Engineering and Computer Science\\
Pennsylvania State University \\
University Park, PA 16801\\
ovm5033@psu.edu, dkifer@cse.psu.edu
}
\maketitle
\begin{abstract}

In this paper, we consider the problem of predicting demographics of geographic units given geotagged Tweets that are composed within these units. Traditional survey methods that offer demographics estimates are usually limited in terms of geographic resolution, geographic boundaries, and time intervals. Thus, it would be highly useful to develop computational methods that can complement traditional survey methods by offering demographics estimates at finer geographic resolutions, with flexible geographic boundaries (i.e. not confined to administrative boundaries), and at different time intervals. While prior work has focused on predicting demographics and health statistics at relatively coarse geographic resolutions such as the county-level or state-level, we introduce an approach to predict demographics at finer geographic resolutions such as the blockgroup-level. For the task of predicting gender and race/ethnicity counts at the blockgroup-level, an approach adapted from prior work to our problem achieves an average correlation of 0.389 (gender) and 0.569 (race) on a held-out test dataset. Our approach outperforms this prior approach with an average correlation of 0.671 (gender) and 0.692 (race). 

\end{abstract}

\section{Introduction}

\noindent Social media data has become increasingly important with applications to many fields such as health \cite{culotta2014estimating} and sociolinguistics \cite{eisenstein2014diffusion}. Furthermore, social media has been used to complement traditional survey methods as a faster and cheaper approach to collect information and make predictions \cite{benton2016collective}.

Collecting demographics data is usually a long and costly process which limits the rate and resolution at which this collection may be performed. The U.S. Census Bureau releases a population census every 10 years with demographics data on multiple geographic resolutions \cite{census2010}. These geographic resolutions follow a hierarchy: each state is divided into counties, then each county is divided into tracts, then each tract is divided into block groups, and each block group is divided into blocks. The census contains demographics data at the block-level and moving up to the state-level. Also, The American Community Survey (ACS), which is a statistical survey of a sample of people prepared by the U.S. Census Bureau, contains 1-year, 3-year, and 5-year moving average estimates of demographics but with lower resolutions (county-level and blockgroup-level) and higher error. Another important issue is that the administrative boundaries change every census, for example, the 2000 block-level boundaries are different from the 2010 block-level boundaries. This raises challenges for researchers who study populations over decades because they need to crosswalk from one census to the next. So, to compare 2000 demographics with 2010 demographics, they need to figure out how would the 2000 demographics look like using boundary definitions from 2010. 

Using geotagged Tweets to predict demographics of geographic units can be a complementary method to survey collection methods as long as their bias can be corrected. This would be a cheaper and faster approach to estimate demographics data at resolutions finer than what traditional sources offer. In addition, this approach is not confined to administrative geographic boundaries but can be adapted to custom geographic boundaries. Also, it is more flexible by permitting collecting demographics data at different time intervals (e.g. 6 months). 

We present a method to predict demographics of high-resolution geographic units using geotagged Tweets. The main idea is to learn to predict demographics of a region based on characteristics of Tweets in that region. An important aspect of our approach is that we do not require labeling of individual Tweets. To evaluate our method, we train models to predict gender and race/ethnicity demographics of Census predefined geographies at different resolutions (block, blockgroup, tract, and county) using 2010 Census demographics data as ground truth. At the block-level, we achieve an average correlation of 0.585 (gender) and 0.487 (race) on a held-out test dataset. We find that our approach significantly improves upon the results of a competing approach adapted from prior work. We also find that for 95\% of blocks, blockgroups, tracts, and counties with at least 100 Twitter users, the relative prediction error is at most 1.98, 1.15, 0.90, and 0.78 respectively.
 
We discuss related work in Section 2, we introduce necessary definitions, notations, and formally define the problem in Section 3, we describe our approach in Section 4, and we present experiments in Section 5. 

\section{Related Work}

\noindent The availability of large-scale geotagged Twitter data has spurred a lot of interest in predicting demographics of geographic units. It is viewed as a cheaper and faster method to draw inferences that can complement traditional survey methods. There are two strategies in general. First is predicting demographics of geographic units directly. Second is predicting demographics of Twitter users in those geographic units individually and then performing some form of aggregation. 

Prior work that employs the first strategy spans several applications. \citeauthor{eisenstein2011discovering} \shortcite{eisenstein2011discovering} predicted demographics of Zip Code Tabulation Areas (ZCTA) using geotagged Tweets and Census data. Similarly, \citeauthor{mohammady2014using} \shortcite{mohammady2014using} predicted race composition of the 100 most populous counties from geotagged Tweets and then used that for individual-level labeling of Twitter users. There is also prior work that focused on health statistics. \citeauthor{schwartz2013characterizing} \shortcite{schwartz2013characterizing} predicted life-satisfaction of counties using county demographics and Tweets. \citeauthor{culotta2014estimating} \shortcite{culotta2014estimating} predicted several health statistics (such as obesity rates) of the 100 most populous counties. \citeauthor{eichstaedt2015psychological} \shortcite{eichstaedt2015psychological} predicted atherosclerotic heart disease mortality rates of counties from Tweets. \citeauthor{ireland2015future} \shortcite{ireland2015future} predicted HIV rates of counties from Tweets. \citeauthor{loff2015predicting} \shortcite{loff2015predicting} predicted well-being of states by estimating the Gallup-Healthways index. \citeauthor{benton2016collective} \shortcite{benton2016collective} used supervised topic models to predict responses to miscellaneous survey questions such as percentage of smokers at the state-level.

As described, these prior approaches made predictions at relatively coarse geographic resolutions such as the county-level which may be due to limitations in ground truth data availability at finer resolutions (e.g. health statistics). Our approach predicts demographics at finer geographic resolutions such as the block-level. Furthermore, we explore how large a population needs to be (i.e. resolution) in order to get accurate predictions.

Related to the second strategy, there is a rich body of work on individual-level prediction tasks focusing on attributes of social media users. These attributes include: gender \cite{burger2011discriminating,bergsma2013broadly,bamman2014gender,vicente2015twitter}, age \cite{schwartz2013personality,moseley2014toward}, ethnicity \cite{chang2010epluribus,chen2015comparative}, income and socio-economic status \cite{preoctiuc2015studying,lampos2016inferring}. In addition, several works predicted a wide-range of attributes including demographics and emotions \cite{culotta2015predicting,volkova2015inferring}. Adapting these methods to solve our problem poses several challenges. For example, it requires supervised training and labeled demographics for large numbers of Twitter users. These labels have to be collected yearly to account for concept drift associated with changing generations. Also, once Twitter users are labeled, it is not immediately obvious how to combine these observations to arrive at the demographics of geographic units since Twitter users with geotagged tweets are a highly biased sample \cite{malik2015population}. To reduce the sampling bias, \citeauthor{culotta2014reducing} \shortcite{culotta2014reducing} weighted the contributions of Twitter users based on their demographics. Also, \citeauthor{almeida2015twitter} \shortcite{almeida2015twitter} sampled a subset of Twitter users in an area according to the distribution of the real population in that area. Both approaches require individual-level labelling of Twitter users. Our method however directly trains to fix the bias. For example, we know the characteristics of Twitter users in an area (biased) and we know the population characteristics of an area, and we learn a function (model) that maps one into the other. 

\section{Preliminaries}

\noindent In this section, we setup some necessary definitions, notations, and formally define the problem. Our dataset is a set of Tweets $\{t_1, t_2, ..., t_m\}$. Each Tweet $t_i$ is a tuple of the form $(loc_i, uid_i, <w^i_1, w^i_2, w^i_3, ...>)$ where $loc$ is the GPS location, $uid$ is the user id, and $<w_1, w_2, w_3, ...>$ is the sequence of tokens in the Tweet. We have a set of geographic units $\{g_1, g_2, ..., g_n\}$. Each $g_i$ is a tuple of the form $(shape_i, \mathbf{y_i})$ where $shape$ is the boundary definition and $\mathbf{y}$ is ground truth count data of a demographic variable (e.g. gender) of the geographic unit. If we have a demographic variable with $k$ mutually exclusive categories, then $\mathbf{y}$ is a $k$ dimensional vector where $y_j$ is the count of category $j$, $j \in [1,k]$ (e.g. for gender, $y_1$ is the count of males and $y_2$ is the count of females). For each Tweet $t_i$, we map it to a geographic unit $g_j$ if $shape_j$ contains $loc_i$. Thus, we group the Tweets into disjoint bags that correspond to the geographic units:
\begin{align*}
    (\{t_1^j\}_{j=1}^{N_1}, \mathbf{y_1}), (\{t_2^j\}_{j=1}^{N_2}, \mathbf{y_2}), ..., (\{t_n^j\}_{j=1}^{N_n}, \mathbf{y_n})
\end{align*}

where geographic unit $i$ has a set of observed Tweets $\{t_i^j\}_{j=1}^{N_i}$ with a total of $N_i$ Tweets and a demographic variable vector $\mathbf{y_i}$. Using such a dataset, our goal is to learn a model $f$ such that for an unseen geographic unit $g$ with a set of Tweets $\{t_g^j\}_{j=1}^{N_g}$, $f(\{t_g^j\}_{j=1}^{N_g}) = \widehat{\mathbf{y_g}}$, where $\widehat{\mathbf{y_g}}$ is an estimate of the true unknown demographics count $\mathbf{y_g}$.

Mathematical notation is defined in table \ref{table:1}.
\begin{table}[t]
\begin{tabular}{|l|l|}
\hline
{\bf Notation} & {\bf Definition}\\
\hline
$W_i$ & the set of tweeted words that appeared in $g_i$.\\
\hline
$c_{w,i}$ & \# of times $w$ appeared in $g_i$.\\
\hline
$C_i$ & \# of total words that appeared in $g_i$.\\
\hline
$u_{w,i}$ & \# of Twitter users who used $w$ in $g_i$.\\
\hline
$U_i$ & \# of Twitter users who tweeted at least \\
& once from $g_i$.\\
\hline
$\mathbf{x_i}$ & Feature vector of $g_i$.\\
\hline
$x_{i,w}$ & Value of feature $w$ in $g_i$ .\\
\hline
$D$ & Dimension of the feature space.\\
\hline
\end{tabular}
\caption{Mathematical notation}
\label{table:1}
\end{table}

\section{Method}

\noindent In this section, we describe our approach to learn the model $f$. Our approach relies on computing a feature vector $\mathbf{x_i} \in \mathbb{R}^D$ given $\{t_i^j\}_{j=1}^{N_i}$, for each geographic unit $i \in [1, n]$. Then, we fit a model to predict $\widehat{\mathbf{y_i}}$ given $\mathbf{x_i}$. Below, we discuss possible feature engineering and modeling choices.

\subsection{Feature Engineering}
In accordance with prior work, we focus on features that are based on lexical content. This is motivated by exploring the most predictive linguistic patterns of demographics. First we discuss possible lexical features, then we discuss possible normalization and transformation schemes that can be applied to these features.

\subsubsection{Features} There are several possible lexical features that can be used to represent geographic units. These include but are not limited to: lexicons, latent topics, words and phrases (bag-of-words), and embeddings. These features are not mutually exclusive and can be combined together.

Lexicons are predefined word-to-category mappings that can be used to represent each geographic unit by the frequency of each category \cite{schwartz2013characterizing,culotta2014estimating}. Lexicons usually have stronger domain assumptions (compared to bag-of-words) \cite{o2011computational} and are limited to specific applications such as health and personality \cite{culotta2014estimating}. So, we do not explore using lexicons as features.

Each geographic unit can also be represented by its distribution over a set of latent topics, most commonly learned using latent Dirichlet allocation \cite{blei2003latent}. \citeauthor{schwartz2013characterizing} \shortcite{schwartz2013characterizing} report that this is better than using lexicons for their task of predicting well-being. \citeauthor{benton2016collective} \shortcite{benton2016collective} explored variants of topic models that are guided by supervision to generate feature representations. Interestingly, they found that the bag-of-words representation is competitive with the best supervised topic models.

We explored using embeddings as features by learning representations of geographic units using Paragraph Vector \cite{le2014distributed}. This is a similar technique to Word2Vec \cite{mikolov2013distributed}, but rather than learning representations of words, it learns representations of paragraphs or documents. We model each $g_i$ as a document consisting of sentences which are the Tweets $\{t_i^j\}_{j=1}^{N_i}$. We found that bag-of-words is competitive with this approach.

Bag-of-words uses words and/or phrases as features instead of using categories or topics. We use this representation because it is simpler and has weaker domain assumptions \cite{o2011computational}. 

\subsubsection{Normalizations} We discuss different ways of counting and normalizing occurrences of words. Our discussion is based on using a bag-of-words representation but these techniques can also be applied to lexicons. We start with computing raw counts of tweeted words, $c_{w,i}$ and $u_{w,i}$, for each word $w$ and region $g_i$ where $i \in [1,n]$. $c_{w,i}$ is the number of times a word $w$ is tweeted in region $g_i$. It is oblivious to the number of Twitter users that used $w$. So, it is possible for the feature vector to be skewed by Twitter users that use a word $w$ many times, either in one or many Tweets. To account for that, $u_{w,i}$ counts the number of distinct Twitter users that used $w$ in $g_i$. Since the distribution of geotagged Tweets and geotag Twitter users is not uniform across regions \cite{malik2015population}, using raw counts such as $c_{w,i}$ and $u_{w,i}$ as feature values will result in highly imbalanced feature vectors for geographic units. To balance these differences between geographic units, we normalize $c_{w,i}$ by dividing by the number of total tweeted words in region $g_i$, $C_i$; and $u_{w,i}$ by dividing by the number of Twitter users that tweeted at least once in region $g_i$, $U_i$. Normalizations help differentiate between geographic units as they take into account the size of total observations. 

Using these mutually exclusive schemes, we compute feature values $v_{i,w}$ for each geographic unit $g_i$ and each word $w$ in the set of tweeted words in $g_i$, $W_i$ (for $w \notin W_i$, $v_{i,w} = 0$):
\begin{itemize}
\item \textbf{Raw Word}: $v_{i,w} = c_{w,i}$.
\item \textbf{Normalized Word:} $v_{i,w} = \frac{c_{w,i}}{C_i}$.
\item \textbf{Raw User:} $v_{i,w} = u_{w,i}$.
\item \textbf{Normalized User:} $v_{i,w} = \frac{u_{w,i}}{U_i}$
\end{itemize}

In our experiments our feature set includes only words that appear in the training split. This is to ensure that we account for the effect of out-of-vocabulary words.

\subsubsection{Transformations} After computing a bag-of-words representation using Raw Word, Raw User, Normalized Word, or Normalized User, we perform feature transformations on the representation. We explore different transformations: Term Frequency-Inverse Document Frequency (TFIDF), Anscombe, Logistic, and Gaussian. Not every transformation is applied to every representation, for example, TFIDF is applied to Raw Word and Raw User, and the rest are applied to Normalized Word and Normalized User. 

If we have a feature vector $\mathbf{v_i}$ for a geographic unit $g_i$ (computed using Raw Word, Raw User, Normalized Word, or Normalized User), we transform it to $\mathbf{x_i}$ by applying an element-wise transformation on each $v_{i,w}$. We apply the transformation only on $v_{i,w}\neq0$ to preserve the sparsity of our feature vectors.

\textbf{TFIDF:} The word distribution across all geographic units has a long-tail shape, with few words appearing in all geographic units and less-frequent words appearing in few geographic units. The motivation behind using TFIDF is to help the model take that into account, and re-weight the word counts inversely. Each geographic unit represents a document, we learn the inverse document frequency in the following manner:
\begin{align*}
    idf(w) &= \log \frac{n}{1 + |\{i \in [1,n]: w \in W_i\}|}
\end{align*}

We use this transformation with Raw Word or Raw User, so $v_{i,w} = c_{w,i}$ or $v_{i,w} = u_{w,i}$. Then, $x_{i,w} = v_{i,w}( idf(w) + 1 )$. Note that we add a 1 to $idf(w)$ because we do not want to completely ignore words that appear in all geographic units. In our experiments, we learn $idf(\cdot)$ based only on the training split. 

\textbf{Anscombe:} We applied this transformation to stabilize the variance of word frequencies. The distribution of a word may be right-skewed (i.e. appears a lot in a few geographic units and appears little elsewhere), the Anscombe transform helps adjust this skewness and make the distribution roughly symmetric. It helps turn a random variable distribution to be more Gaussian \cite{anscombe1948transformation}. \citeauthor{schwartz2013personality} \shortcite{schwartz2013personality} applied this transformation in predicting select demographics of Facebook users. We use this transformation with Normalized Word or Normalized User, so $v_{i,w} = \frac{c_{w,i}}{C_i}$ or $v_{i,w} = \frac{u_{w,i}}{U_i}$. Then, $x_{i,w} = 2\sqrt{v_{i,w} + \frac{3}{8}} $.

\textbf{Logistic and Gaussian:} Inspired by the use of activation functions in neural networks, we explored applying a non-linear activation function $\phi(\cdot)$ on our word frequencies (computed with Normalized Word or Normalized User). Our intuition is that the non-linearity of $\phi(\cdot)$ would help increase the capacity of the model. We separately used $\phi(x) = \frac{1}{1 + e^{-x}}$ (Logistic) and $\phi(x) = e^{-{x}^2}$ (Gaussian). We set $x_{i,w} = \phi(v_{i,w})$, where $v_{i,w} = \frac{c_{w,i}}{C_i}$ or $v_{i,w} = \frac{u_{w,i}}{U_i}$. We also explored other activation functions such as TanH, ArcTan, and Softsign but they did not show promising results.

\subsection{Modeling}
We explore two variants of the problem: predicting demographics of geographic units when population size is unknown and when population size is known.

\subsubsection{Population Size is Unknown} In this setting we would like to predict demographic counts (e.g. gender) $y_{i,j}$ of a region $g_i$ for each category (e.g. male and female) $j \in [1, k]$ without access to the population size of $g_i$. We choose a linear regression model for scalability. For each category $j \in [1,k]$, we optimize the following objective function:
\begin{align*}
    \mathbf{w}_j &= \arg\min\limits_{\mathbf{w}_j} \frac{1}{2n} \sum_{i=1}^{n} (\mathbf{w}_j\cdot\mathbf{x_i} - y_{i,j})^2 + \lambda||\mathbf{w}_j||_2^2
\end{align*}

where $\mathbf{w}_j$ is the weight vector learned for category $j$, and $\lambda$ is an $l2$ regularization parameter to prevent overfitting. We also explored $l1$ and $ElasticNet$ regularizations, but they yielded similar results.

For a region $g_u$ with an unknown demographic category count $y_{u,j}$, we map the bag of Tweets $\{t_u^j\}_{j=1}^{N_u}$ in region $g_u$ to a transformed feature vector $\mathbf{x_u}$ (we use the same configuration that is used in optimizing the objective function, e.g. Raw Word with TFIDF) and then estimate $y_{u,j}$, where $\hat{y_{u,j}} = \mathbf{w}_j\cdot\mathbf{x_u}$.

\subsubsection{Population Size is Known} In this setting we assume that we have access to the true population count $p_i$ in region $g_i$. We fit a linear regression model to predict $\log \frac{y_{i,j}}{y_{i,q}}$ which is log of the ratio of a demographic category count $y_{i,j}$ to another demographic category count $y_{i,q}$. We choose one demographic category as the denominator (e.g. $q=1$) and then learn $\mathbf{w}_j$ for $j \in [2,k]$ by optimizing the following objective function:
\begin{align*}
    \mathbf{w}_j &= \arg\min\limits_{\mathbf{w}_j} \frac{1}{2n} \sum_{i=1}^{n} (\mathbf{w}_j\cdot\mathbf{x_i} - \log \frac{y_{i,j}}{y_{i,q}} )^2 + \lambda||\mathbf{w}_j||_2^2
\end{align*}

To estimate a demographic category count $y_{i,j}$ for region $g_i$ using $p_i$, we compute:

\[
\hat{y_{i,j}} = \left\{
        \begin{array}{ll}
            \frac{1}{1 + \sum_{m=2}^{k} e^{\mathbf{w}_m\cdot\mathbf{x_i}}} p_i & \quad j=q \\[1em]
            \frac{e^{\mathbf{w}_j\cdot\mathbf{x_i}}}{1 + \sum_{m=2}^{k} e^{\mathbf{w}_m\cdot\mathbf{x_i} }} p_i & \quad j\neq q
        \end{array}
    \right.,
\]

\section{Experiments}

We evaluate our approach and competing approaches (baselines) on both variants of the problem using Census predefined geographies at different resolutions: block, block group, tract, and county. Among the four, block-level is the highest resolution, and county-level is the lowest resolution. In the following subsections we provide details about our experiments: baselines, data, preprocessing, training, and results. 

\subsection{Baselines}
We compare our approach with an approach adapted from \cite{mohammady2014using}, where they used Tweets to predict race/ethnicity composition of counties. In their approach, they used a bag-of-words representation normalized by Twitter users with tweeted words and words from description fields of Twitter users as features. We adapt their approach by using the same types of features. We compute a bag-of-words representation using User Normalization, and then we train a model with this representation and evaluate it on both variants of the problem. Note that in our competing configurations we do not use features from the description field of Twitter users.

In the setting where population size is known, we also compare our models with a baseline that always uses gender and race/ethnicity proportions at the national level to predict category counts of blocks, blockgroups, tracts, and counties. The 2015 national level estimates of proportions are: Male (49.2\%), Female (50.8\%), White (61.6\%), Black or African American (12.4\%), Asian (5.4\%), Hispanic or Latino (17.6\%), Other (3\%) \cite{census2016}.

\subsection{Data}

We collected a large dataset of geotagged Tweets using Twitter's Streaming API from June 12, 2013 to January 31, 2014. We only included Tweets composed in the contiguous U.S. which consists of the 48 adjoining states and Washington D.C. and does not include Alaska and Hawaii for example. We used a bounding box of $[125.0011,66.9326]W \times [24.9493,49.5904]N$. 

Based on a Tweet's GPS coordinates, we annotate it with the geographic identifier (GEOID) of the block that it appeared in. The U.S. Census Bureau provides geographic boundary files (shapefiles) for each state, where each shapefile contains the boundary definitions for all the blocks in that state. This enables us to map each Tweet to its respective block. We used 2010 shapefile definitions to match with 2010 Census demographics data. Overall, we had 565,350,007 Tweets annotated with block-level GEOIDs. 

\subsubsection{Demographics Data} We used data from the 2010 Census. The U.S. Census Bureau provides aggregate count data on different demographics such as gender, age, and race at multiple geographic levels (including block-level to county-level). We specifically used data from the Summary File 1 tables P12 and P5, for gender and race/ethnicity, respectively. We used the Data Finder tool provided by the National Historical Geographic Information System \cite{data2010} to collect this data. We collected gender and race/ethnicity data at the block-level up to the county-level. For gender we used two categories: Male and Female. For race/ethnicity we used five categories: White, African American or Black, Asian, Hispanic, and Other. In each case, the categories are mutually exclusive. Note that there is a time difference of two and a half years between the demographics data and Twitter data. This can bias results and is a direction for future work. 

\subsection{Preprocessing}

Twitter is filled with spam and organizational accounts that post content we deem irrelevant to our application, as we are interested in content produced by personal accounts. To reduce the likelihood of including content from organizational/spam accounts, we removed Tweets from accounts with more than 1000 followers or 1000 friends \cite{lee2011seven,mccorriston2015organizations} and Tweets containing URLs \cite{guo2014detecting}. We also removed Retweets by checking for the existence of \texttt{retweeted\_status} field or the \texttt{RT} token in Tweet text itself. Consequently, our dataset got narrowed down to 423,622,202 Tweets with 4,027,594 unique Twitter users.

To build a bag-of-words representation, we split the text of the Tweets into unigram tokens. There are several things to consider when tokenizing Tweets such as: hashtags, username mentions, emails, html entities, emoticons, etc. For this task, we used \texttt{Twokenize} \cite{tokenizepy}, a tokenizer designed for Tweets which treats hashtags, emoticons, blocks of punctuation marks, and other symbols as tokens. After tokenizing, we removed username mentions, emails, single punctuation marks, and English stopwords. We converted all tokens to lower case. We also chose to keep emoticons as \citeauthor{o2010mixture} (2010) showed in their analysis that groups with certain demographics (high percentage of Hispanics) use emoticons a lot. 

\subsection{Training}

For a given geographic resolution, we randomly split the geographic units that have Tweets into 90\% training and 10\% testing (e.g. we train on 90\% of blocks, and predict  demographics of the remaining 10\%). 10\% of the geographic units in the training split were chosen randomly as a validation set. Note that this splitting is done separately for each geographic resolution. We have $n_{train}$ - $n_{test}$ examples: 5,188,608 - 576,513 (block); 194,610 - 21,624 (blockgroup); 65,239 - 7,249 (tract); 2,798 - 311 (county). Then we compute a configuration (e.g. Raw User with TFIDF) and use all the words that appear in the training split as features (more than 22 million features). Since this is a large-scale learning problem we utilize stochastic gradient descent (SGD) to fit our models. To use SGD, we have to choose a learning rate update policy. We used inverse scaling:
\begin{align*}
    \eta^{\tau} &= \frac{\eta_0}{\tau^\rho}
\end{align*}

Where $\eta_0$ is the initial learning rate, $\tau$ is the time step (indexed by epoch and training example), and $\rho$ is a hyper-parameter that affects the decrease rate of the learning rate. There are several hyper-parameters that need to be selected before training. We performed a grid search on the following hyper-parameters:
\begin{itemize}
\item $\lambda \in \{10^{-6}, 10^{-5}, 10^{-4},0.001,0.01, 0.1\}$
\item $\eta_0 \in \{10^{-6}, 10^{-5}, 10^{-4},0.001,0.01, 0.1,1.0,10.0\}$
\end{itemize}

The hyper-parameter combination that scored the highest $R^2$ (coefficient of determination) score on the validation set was used for training on the entire training split (training and validation). We shuffled the training dataset after each epoch of SGD training. We fixed the number of epochs to 10 and $\rho=0.25$. Early experiments showed that training for more epochs (e.g. 100) does not improve the performance significantly, and likewise changing the value of $\rho$.  


\sisetup{detect-weight=true,detect-inline-weight=math}
\sisetup{detect-weight=true,detect-inline-weight=math}
\begin{table*}[t]
\centering
\begin{tabular}{|l|l|l|l|S[table-format=1.3]|S[table-format=1.3]|S[table-format=1.3]|S[table-format=1.3]|S[table-format=1.3]|S[table-format=1.3]|S[table-format=1.3]|S[table-format=1.3]|S[table-format=1.3]|}
\hline
                             &                               &                             & \multicolumn{4}{l|}{\bfseries{Population Size Unkown}}                                                               & \multicolumn{6}{l|}{\bfseries{Population Size Known}}                                                                                                                      \\ \cline{4-13} 
\multirow{-2}{*}{\bfseries{Res}} & \multirow{-2}{*}{\bfseries{Demo}} & \multirow{-2}{*}{\bfseries{Metric}}    & B                        & WL                       & UA                       & UG                       & C                        & B                        & WA                       & WL                       & UG                       & RUT                       \\ \hline
                             &                               & \cellcolor[HTML]{C0C0C0}$r$  & \cellcolor[HTML]{C0C0C0} 0.183 & \cellcolor[HTML]{C0C0C0} 0.576& \cellcolor[HTML]{C0C0C0} 0.554& \cellcolor[HTML]{C0C0C0} \bfseries{0.585} & \cellcolor[HTML]{C0C0C0} 0.957 & \cellcolor[HTML]{C0C0C0} 0.958& \cellcolor[HTML]{C0C0C0} 0.960& \cellcolor[HTML]{C0C0C0} \bfseries{0.960}& \cellcolor[HTML]{C0C0C0} 0.959& \cellcolor[HTML]{C0C0C0} \bfseries{0.959}\\ \cline{3-13} 
                             & \multirow{-2}{*}{Gender}      & $R^2$                        & 0.033 & 0.333& 0.308& \bfseries{0.344} & 0.922 & 0.914& 0.916& \bfseries{0.917}& 0.916& \bfseries{0.915}\\ \cline{2-13} 
                             &                               & \cellcolor[HTML]{C0C0C0}$r$   & \cellcolor[HTML]{C0C0C0} 0.195 & \cellcolor[HTML]{C0C0C0} 0.480& \cellcolor[HTML]{C0C0C0} 0.462& \cellcolor[HTML]{C0C0C0} \bfseries{0.487} & \cellcolor[HTML]{C0C0C0} 0.620& \cellcolor[HTML]{C0C0C0} 0.708& \cellcolor[HTML]{C0C0C0} 0.701& \cellcolor[HTML]{C0C0C0} 0.700& \cellcolor[HTML]{C0C0C0} 0.704& \cellcolor[HTML]{C0C0C0} \bfseries{0.714}\\ \cline{3-13} 
\multirow{-4}{*}{Block}      & \multirow{-2}{*}{Race}        & $R^2$                          & 0.039 & 0.235& 0.218& \bfseries{0.243} & 0.377 & 0.216& 0.320& 0.376& 0.391& \bfseries{0.449}  \\ \hline
                             &                               & \cellcolor[HTML]{C0C0C0}$r$   & \cellcolor[HTML]{C0C0C0} 0.389 & \cellcolor[HTML]{C0C0C0} \bfseries{0.671}& \cellcolor[HTML]{C0C0C0} 0.670& \cellcolor[HTML]{C0C0C0} 0.667 & \cellcolor[HTML]{C0C0C0} 0.980 & \cellcolor[HTML]{C0C0C0} 0.982& \cellcolor[HTML]{C0C0C0} \bfseries{0.982}& \cellcolor[HTML]{C0C0C0} 0.982& \cellcolor[HTML]{C0C0C0} 0.979& \cellcolor[HTML]{C0C0C0} \bfseries{0.982}\\ \cline{3-13} 
                             & \multirow{-2}{*}{Gender}      & $R^2$                          & 0.150 & \bfseries{0.449}& 0.448& 0.444 & 0.959 & 0.963& \bfseries{0.964}& 0.964& 0.958& \bfseries{0.964} \\ \cline{2-13} 
                             &                               & \cellcolor[HTML]{C0C0C0}$r$   & \cellcolor[HTML]{C0C0C0} 0.569 & \cellcolor[HTML]{C0C0C0} \bfseries{0.692}& \cellcolor[HTML]{C0C0C0} 0.690& \cellcolor[HTML]{C0C0C0} 0.683 & \cellcolor[HTML]{C0C0C0} 0.407 & \cellcolor[HTML]{C0C0C0} 0.780& \cellcolor[HTML]{C0C0C0} 0.783& \cellcolor[HTML]{C0C0C0} 0.788& \cellcolor[HTML]{C0C0C0} 0.783& \cellcolor[HTML]{C0C0C0} \bfseries{0.835}\\ \cline{3-13} 
\multirow{-4}{*}{BG} & \multirow{-2}{*}{Race}        & $R^2$                         & 0.311 & \bfseries{0.491}& 0.489& 0.480 &  0.184 & 0.613& 0.632& 0.632& 0.617& \bfseries{0.676} \\ \hline
                             &                               & \cellcolor[HTML]{C0C0C0}$r$   & \cellcolor[HTML]{C0C0C0} 0.458 & \cellcolor[HTML]{C0C0C0} 0.723& \cellcolor[HTML]{C0C0C0} \bfseries{0.726}& \cellcolor[HTML]{C0C0C0} 0.723 & \cellcolor[HTML]{C0C0C0} 0.985 & \cellcolor[HTML]{C0C0C0} 0.986& \cellcolor[HTML]{C0C0C0} \bfseries{0.987}& \cellcolor[HTML]{C0C0C0} \bfseries{0.987}& \cellcolor[HTML]{C0C0C0} \bfseries{0.982}& \cellcolor[HTML]{C0C0C0} \bfseries{0.988}\\ \cline{3-13} 
                             & \multirow{-2}{*}{Gender}      & $R^2$                          & 0.204 & 0.523& \bfseries{0.523}& 0.522 & 0.965 & 0.972& \bfseries{0.973}& \bfseries{0.972}& \bfseries{0.965}& \bfseries{0.974}  \\ \cline{2-13} 
                             &                               & \cellcolor[HTML]{C0C0C0}$r$  & \cellcolor[HTML]{C0C0C0} 0.669 & \cellcolor[HTML]{C0C0C0} 0.756& \cellcolor[HTML]{C0C0C0} \bfseries{0.757}& \cellcolor[HTML]{C0C0C0} 0.752 & \cellcolor[HTML]{C0C0C0} 0.365& \cellcolor[HTML]{C0C0C0} 0.796& \cellcolor[HTML]{C0C0C0} 0.833& \cellcolor[HTML]{C0C0C0} 0.838& \cellcolor[HTML]{C0C0C0} 0.825& \cellcolor[HTML]{C0C0C0} \bfseries{0.884}\\ \cline{3-13} 
\multirow{-4}{*}{Tract}      & \multirow{-2}{*}{Race}        & $R^2$                         & 0.457 & 0.585& \bfseries{0.587}& 0.578 & 0.151& 0.640& 0.705& 0.709& 0.696& \bfseries{0.761}\\ \hline
                             &                               & \cellcolor[HTML]{C0C0C0}$r$ & \cellcolor[HTML]{C0C0C0} 0.809 & \cellcolor[HTML]{C0C0C0} 0.986& \cellcolor[HTML]{C0C0C0} 0.985& \cellcolor[HTML]{C0C0C0} \bfseries{0.988} & \cellcolor[HTML]{C0C0C0} 0.999 & \cellcolor[HTML]{C0C0C0} 0.999& \cellcolor[HTML]{C0C0C0} 0.999& \cellcolor[HTML]{C0C0C0} 0.999& \cellcolor[HTML]{C0C0C0} 0.999& \cellcolor[HTML]{C0C0C0} \bfseries{0.999}\\ \cline{3-13} 
                             & \multirow{-2}{*}{Gender}      & $R^2$                        & 0.513 & 0.969& 0.968& \bfseries{0.973} & 0.999 & 0.999& 0.999& 0.999& 0.999& \bfseries{0.999}\\ \cline{2-13} 
                             &                               & \cellcolor[HTML]{C0C0C0}$r$    & \cellcolor[HTML]{C0C0C0} 0.745 & \cellcolor[HTML]{C0C0C0} 0.879& \cellcolor[HTML]{C0C0C0} 0.868& \cellcolor[HTML]{C0C0C0} \bfseries{0.898} & \cellcolor[HTML]{C0C0C0} 0.850 & \cellcolor[HTML]{C0C0C0} 0.963& \cellcolor[HTML]{C0C0C0} 0.834& \cellcolor[HTML]{C0C0C0} 0.945& \cellcolor[HTML]{C0C0C0} \bfseries{0.962}& \cellcolor[HTML]{C0C0C0} 0.956\\ \cline{3-13} 
\multirow{-4}{*}{County}     & \multirow{-2}{*}{Race}        & $R^2$                         & 0.491 & 0.740& 0.726& \bfseries{0.775} & 0.694 & 0.826& 0.551& 0.837& \bfseries{0.886}& 0.796\\ \hline
\multicolumn{13}{c}{A - Anscombe, B - \cite{mohammady2014using}, BG - Blockgroup, C - National Proportions, Demo - Demographic}\\
\multicolumn{13}{c}{G - Gaussian, L - Logistic, Res - Resolution, RU - Raw User, T - TFIDF, U - Normalized User, W - Normalized Word}\\
\end{tabular}
\caption{Performance results on multiple resolutions across gender and race/ethnicity prediction tasks. In each problem variant, bold results in a row represent configurations that are statistically indistinguishable using a paired t-test with $p$-value $\geq 0.05$.}
\label{table:2}
\end{table*}

\subsection{Results}
We evaluate the baselines and our models (with different feature engineering configurations) on both variants of the problem (population count is unknown vs. known). In both variants we predict demographic category counts ($y_j$ for $j \in[1,k]$) and evaluate the performance on the test split using Pearson correlation $r$ and the coefficient of determination $R^2$. Note that $R^2$ compares the performance of a model relative to the baseline of always predicting the average value of the test set. Table \ref{table:2} summarizes the results of our experiments (due to limited space we include only results of our best configurations). For Pearson correlations, all of them are statistically significant using a two-tailed test with $p$-value $< 10^{-4}$. 

\subsubsection{Population Size is Unknown} In this setting, we find that our models outperform the baseline adapted from \cite{mohammady2014using}, where correlation $r$ (averaged across gender and race) improves by a factor of 2.84 (block), 1.41 (blockgroup), 1.31 (tract), and 1.21 (county) using our User Normalization with Gaussian configuration.

We find that our feature transformations (e.g. Anscombe) improve upon the results of plain bag-of-words representations significantly. Compared with plain User Normalization (which is better than plain Word Normalization), correlation $r$ improves by a factor of 1.88 averaged across gender and race and then across resolutions using User Normalization with Gaussian. This tells us that such transformations help predict demographics better. 

\subsubsection{Population Size is Known} In this setting, we find that predictions improve overall and are better than predictions in the other variant (this is expected because we know total population size). For the task of predicting gender counts (in this case $q$ corresponds to female), the baselines and our models perform comparably to each other. This is the case because there is little variation in gender proportions across geographies, so if total population is known, even a baseline that predicts half of that for both categories will do well.

For the task of predicting race counts (in this case $q$ corresponds to white), we find that our models outperform the baseline of predicting counts based on national-level proportions. Correlation $r$ improves by a factor of 1.15 (block), 2.05 (blockgroup), 2.42 (tract), and 1.12 (county) using our Raw User with TFIDF configuration. Compared to the baseline adapted from \cite{mohammady2014using}, correlation $r$ improves by a factor of 1.01 (block), 1.07 (block group), 1.11 (tract), and 1.05 (county). 
 
We have shown that when we are interested in learning proportions of demographics, our approach outperforms existing baselines. Interestingly, we find that our best configuration is Raw User with TFIDF (which is not the case with the other variant). This may in part be due to the fact that User Normalization reduces the skewness of feature vectors and the fact that the TFIDF transformation increases the importance of less-frequent words and dampens the importance of more frequent words.

\begin{figure}
    \includegraphics{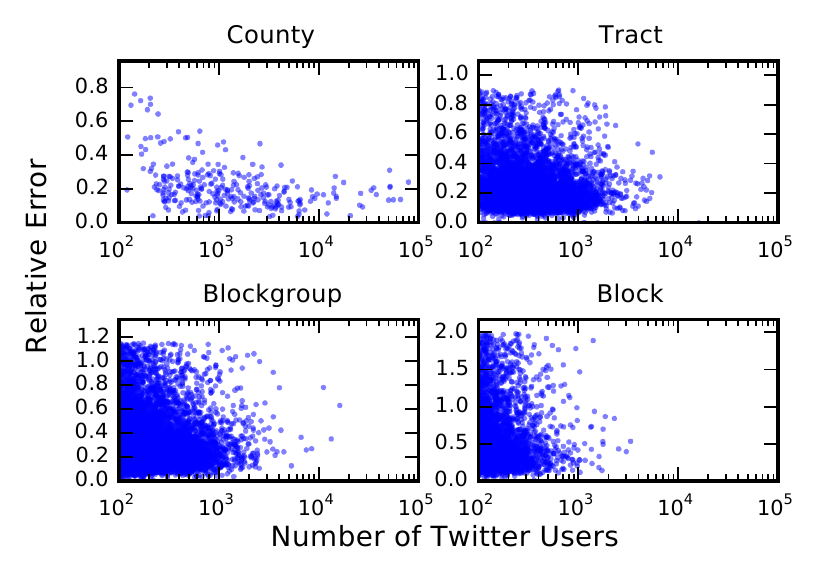}
    \caption{Relative error for 95\% of geographic regions with at least 100 Twitter users (Raw User with TFIDF).}
    \label{figure:3}
\end{figure}

\subsubsection{Prediction Accuracy vs. Number of Twitter Users} We explore the finest geographic resolution that we can predict demographics at, with reasonable accuracy. We plot the average relative error across gender and race versus number of Twitter users (those with geotagged Tweets) in Figure \ref{figure:3}. We find that 95\% of geographic regions with at least 100 Twitter users, have low relative errors. In these regions, the relative error is at most 1.98 (block), 1.15 (blockgroup), 0.90 (tract), and 0.78 (county).

\section{Conclusion}

In this paper, we have shown that geotagged Tweets can be used to estimate demographics of high-resolution geographies. Our method can be used as an alternative or a complement to survey methods. We have shown that certain feature transformations such as Anscombe, TFIDF, Logistic, and Gaussian significantly improve prediction performance relative to competing baselines. We have also shown that the our method is able to learn proportions of demographic categories and can provide accurate predictions at regions with at least 100 Twitter users. 

For future work, it is worth bringing attention to the effect of data sampling rate on prediction. According to \citeauthor{eisenstein2014diffusion} \shortcite{eisenstein2014diffusion}, word frequencies normalized by users (Normalized User) are not invariant to the sampling rate of the data. If we remove half the tweets, then these frequencies will decrease because the number of users will decrease more slowly than raw word counts. So, it would be interesting to investigate methods that can minimize the variance of such normalizations to the sampling rate.

\section{Acknowledgments}
This work was supported by NSF award \#1054389. We also thank Guangqing Chi for providing the Twitter data.


\fontsize{9.0pt}{10.0pt}
\selectfont
\bibliography{Biblio-Database}
\bibliographystyle{aaai}

\end{document}